\documentclass[10pt,twocolumn,letterpaper]{article}
\usepackage[pagenumbers]{cvpr}

\usepackage[normalem]{ulem}
\usepackage{bm}
\usepackage{multirow} 
\usepackage{xspace}
\usepackage{enumitem}

\definecolor{myPink}{RGB}{255,105,180}
\definecolor{cvprblue}{rgb}{0.21,0.49,0.74}
\usepackage[pagebackref,breaklinks,colorlinks,allcolors=cvprblue,urlcolor=myPink]{hyperref}

\newcommand{\methodname}{Mesh4D\xspace}
\newcommand{\method}{Mesh4D\xspace}

\newcommand{\dlt}[1]{{\color{CornflowerBlue}#1}}

\newcommand{\bone}{\boldsymbol{b}}
\newcommand{\decoder}{\mathcal{D}}
\newcommand{\deformation}{\mathcal{T}}
\newcommand{\faces}{\mathcal{F}}
\newcommand{\features}{\boldsymbol{h}}
\newcommand{\image}{\boldsymbol{I}}
\newcommand{\latent}{\boldsymbol{z}}
\newcommand{\mesh}{\mathcal{M}}
\newcommand{\noise}{\boldsymbol{\epsilon}}
\newcommand{\normal}{\boldsymbol{n}}
\newcommand{\parameters}{\boldsymbol{\theta}}
\newcommand{\positionalEncoding}{\operatorname{PE}}
\newcommand{\queryPoints}{\mathcal{Q}}
\newcommand{\real}{\mathbb{R}}
\newcommand{\skinning}{\boldsymbol{w}}
\newcommand{\sparsePoints}{\boldsymbol{h}^{\prime\prime}}
\newcommand{\velocity}{\boldsymbol{v}}
\newcommand{\vertices}{\mathcal{V}}
\newcommand{\video}{\mathcal{I}}
\newcommand{\spatialemb}{\boldsymbol{p}}

\makeatletter
\renewcommand{\paragraph}{%
  \@startsection{paragraph}{4}%
  {\z@}{-0.5em}{-0.5em}%
  {\normalfont\normalsize\bfseries}%
}
\makeatother

\def\paperID{*****} 
\def\confName{CVPR}
\def\confYear{2026}

\title{\method: 4D Mesh Reconstruction and Tracking from Monocular Video}

\newcommand\rurl[1]{%
  \href{https://#1}{\nolinkurl{#1}}%
}

\author{Zeren Jiang$^{1}$
\quad
Chuanxia Zheng$^{1,2}$
\quad
Iro Laina$^{1}$
\quad
Diane Larlus$^{3}$
\quad
Andrea Vedaldi$^{1}$ \\
 $^1$VGG, University of Oxford \quad
 $^2$Nanyang Technological University \quad
 $^3$Naver Labs Europe \\
{\tt\small \{zeren, cxzheng, iro, vedaldi\}@robots.ox.ac.uk \quad diane.larlus@naverlabs.com}\\[0.1em]
\small\rurl{mesh-4d.github.io}
}

\begin{document}
\twocolumn[{
\maketitle
\vspace{-1em}
\begin{center}
\includegraphics[width=\linewidth]{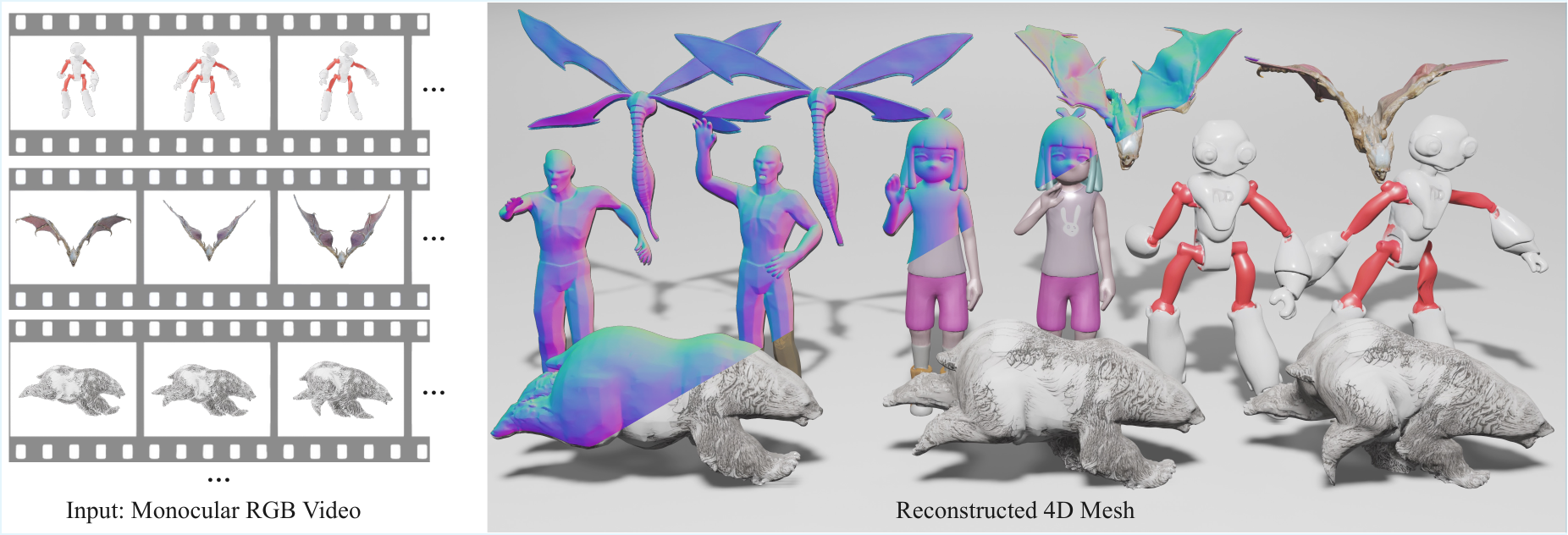}
\end{center}
\vspace{-1em}
\captionof{figure}{
\textbf{Illustration of \method}.
Given a monocular RGB video as input, \method generates a complete animated 3D mesh and its deformation. Each 4D reconstruction is shown at several time steps, the top layer displaying normals and the bottom one textured meshes.}%
\label{fig:splash}
\vspace{3em}
}]

\newcommand{\tableRecon}{
\begin{table}[tb!]
\centering
\resizebox{\linewidth}{!}{%
\begin{tabular}{@{}l cccc@{}}
\toprule
\multirow{2}*{Method} & \multicolumn{3}{c}{\textbf{Reconstruction}} & \multicolumn{1}{c}{\textbf{Tracking}} \\ 
\cmidrule(lr){2-4}
\cmidrule(lr){5-5}
~  & IoU~$\uparrow$ & P2S~$\downarrow$  & Chamfer~$\downarrow$ & $\ell_2$-Corr~$\downarrow$  \\ 
\midrule
HY3D 2.1~\cite{hunyuan3d2025hunyuan3d21} & 0.3071 & 0.0376 & 0.0370 & N/A \\

L4GM~\cite{ren2024l4gm} & N/A & 0.0459 & 0.0505 & N/A \\

GVFD~\cite{zhang2025GVFD} & N/A & 0.0345 & 0.0378 & 0.0514 \\
Ours  & \underline{0.3731} & \underline{0.0287} & \underline{0.0273} & \underline{0.0384} \\
Ours (Aligned)  & \textbf{0.3949} & \textbf{0.0261} & \textbf{0.0243} & \textbf{0.0338} \\
\bottomrule
\end{tabular}
}
\caption{\textbf{Quantitative evaluation} for geometry and tracking on our proposed benchmark, a subset of Objaverse.
All instantiations of our model outperform previous state-of-the-art models.
3D-GS based methods do not explicitly define inner or outer surface, so it is not applicable for volumatric IoU evaluation.
Besides, HY3D~\cite{hunyuan3d2025hunyuan3d21} and L4GM~\cite{ren2024l4gm} predict independent mesh or points per-frame, which do not support tracking evaluation directly.}
\label{tab:recon}
\vspace{-0.60cm}
\end{table}
}

\newcommand{\tableNVS}{
\begin{table}[tb!]
\centering
\resizebox{\linewidth}{!}{%
\begin{tabular}{@{}l cccc c@{}}
\toprule
\multirow{2}*{Method} & \multicolumn{4}{c}{\textbf{Per-frame Quality}} & \multicolumn{1}{c}{\textbf{Consistency}} \\ 
\cmidrule(lr){2-5}
\cmidrule(lr){6-6}
~  & PSNR~$\uparrow$ & SSIM~$\uparrow$ & LPIPS~$\downarrow$ & CLIP~$\uparrow$ & FVD~$\downarrow$ \\ 
\midrule
HY3D 2.1~\cite{hunyuan3d2025hunyuan3d21} & 19.14 & 0.8976 & 0.1195 & \textbf{0.9174} & 692.2 \\

L4GM~\cite{ren2024l4gm} & 18.07 & 0.8939 & 0.1453 & 0.8954 & 747.3 \\

GVFD~\cite{zhang2025GVFD} & 17.31 & 0.8912 & 0.1459 & 0.8802 & 905.0 \\

Ours  & \underline{19.67} & \underline{0.9018} & \underline{0.1087} & \underline{0.9141} & \underline{601.9} \\

Ours (Aligned)  & \textbf{19.88} & \textbf{0.9030} & \textbf{0.1052} & 0.9141 & \textbf{572.7} \\
\bottomrule
\end{tabular}
}
\caption{\textbf{Quantitative evaluation} for novel view synthesis on our proposed benchmark, a subset of Objaverse. We achieve the best performance on both frame-wise quality and video consistency.}
\label{tab:nvs}
\end{table}
}

\newcommand{\tableAblation}{
\begin{table}[tb!]
\centering
\resizebox{\linewidth}{!}{%
\begin{tabular}{@{}l cccc@{}}
\toprule
\multirow{2}*{Method} & \multicolumn{3}{c}{\textbf{Reconstruction}} & \multicolumn{1}{c}{\textbf{Tracking}} \\ 
\cmidrule(lr){2-4}
\cmidrule(lr){5-5}
~  & IoU~$\uparrow$ & P2S~$\downarrow$  & Chamfer~$\downarrow$ & $\ell_2$-Corr~$\downarrow$  \\ 
\midrule
w/o temp\&global attention & 0.6328 & 0.0153 & 0.0113 & 0.0160 \\

w/o skeleton information & 0.6704 & 0.0148 & 0.0107 & 0.0138 \\

Ours  & \textbf{0.7039} & \textbf{0.0144} & \textbf{0.0099} & \textbf{0.0117} \\
\bottomrule
\end{tabular}
}
\caption{\textbf{Quantitative ablation study} for our deformation VAE on our proposed benchmark, a subset of Objaverse. We demonstrate the effectiveness of our key designs.}
\label{tab:ablation}
\vspace{-0.60cm}
\end{table}
}

\newcommand{\tablePretrain}{
\begin{table}[tb!]
\centering
\resizebox{\linewidth}{!}{%
\begin{tabular}{@{}l cccc@{}}
\toprule
\multirow{2}*{Method} & \multicolumn{3}{c}{\textbf{Reconstruction}} & \multicolumn{1}{c}{\textbf{Tracking}} \\ 
\cmidrule(lr){2-4}
\cmidrule(lr){5-5}
~  & IoU~$\uparrow$ & P2S~$\downarrow$  & Chamfer~$\downarrow$ & $\ell_2$-Corr~$\downarrow$  \\ 
\midrule
w/o pretrained & 0.0819 & 0.0954 & 0.0854 & 0.2063 \\

w pretrained  & \textbf{0.3433} & \textbf{0.0327} & \textbf{0.0308} & \textbf{0.0601} \\
\bottomrule
\end{tabular}
}
\caption{\textbf{Quantitative evaluation} for using pretrained weights.}
\label{tab:pretrained}
\vspace{-0.50cm}
\end{table}
}

\newcommand{\tableCFG}{
\begin{table}[tb!]
\centering
\resizebox{\linewidth}{!}{%
\begin{tabular}{@{}l cccc@{}}
\toprule
\multirow{2}*{Method} & \multicolumn{3}{c}{\textbf{Reconstruction}} & \multicolumn{1}{c}{\textbf{Tracking}} \\ 
\cmidrule(lr){2-4}
\cmidrule(lr){5-5}
~  & IoU~$\uparrow$ & P2S~$\downarrow$  & Chamfer~$\downarrow$ & $\ell_2$-Corr~$\downarrow$  \\ 
\midrule

w CFG  & 0.3949 & 0.0261 & 0.0243 & 0.0338 \\

w/o CFG & \textbf{0.3973} & \textbf{0.0258} & \textbf{0.0238} & \textbf{0.0335} \\
\bottomrule
\end{tabular}
}
\caption{\textbf{Ablation study for classifier-free guidance (CFG).} The one without using CFG get slightly better results.}
\label{tab:cfg}
\vspace{-0.30cm}
\end{table}
}

\begin{abstract}
We propose \method, a feed-forward model for monocular 4D mesh reconstruction.
Given a monocular video of a dynamic object, our model reconstructs the object's complete 3D shape and motion, represented as a deformation field.
Our key contribution is a compact latent space that encodes the entire animation sequence in a single pass.
This latent space is learned by an autoencoder that, during training, is guided by the skeletal structure of the training objects, providing strong priors on plausible deformations.
Crucially, skeletal information is not required at inference time.
The encoder employs spatio-temporal attention, yielding a more stable representation of the object's overall deformation.
Building on this representation, we train a latent diffusion model that, conditioned on the input video and the mesh reconstructed from the first frame, predicts the full animation in one shot.
We evaluate \method on reconstruction and novel view synthesis benchmarks, outperforming prior methods in recovering accurate 3D shape and deformation.
\end{abstract}
    
\section{Introduction}%
\label{sec:intro}

In \textit{monocular 4D mesh reconstruction}, we are interested in reconstructing the complete 3D shape and motion of dynamic objects from monocular RGB videos.
Solving this problem has many applications in computer vision, graphics, and robotics.
Automating this pipeline is particularly appealing, as manual 3D modeling and animation are costly, time-consuming, and require domain expertise.
However, this is also very challenging because monocular videos only show parts of the objects.
Since the goal is to recover the objects in full, one must complete their 3D shape and track their deformation throughout the video.

Often~\cite{zollhofer2014real,newcombe2015dynamicfusion,innmann2016volumedeform} this problem has been tackled using analysis by synthesis, where an animated 3D mesh is optimized iteratively to fit the input RGB(D) frames. %
However, these methods can only reconstruct the \emph{visible parts} of the object and often produce incomplete or noisy results due to occlusions and sensor noise.
Recent feed-forward methods~\cite{sucar2025dynamic,st4rtrack2025,jin2024stereo4d} have emerged as promising alternatives.
They learn to reconstruct dynamic 3D geometry from RGB videos in a single pass.
Even so, these methods focus on reconstructing the visible geometry and only track dense correspondences across pairs of frames, rather than across the whole sequence.
As a result, they often fail to capture the entire 4D structure of animated objects.

Reconstructing and tracking the parts of the geometry that are not visible in the video requires strong 3D and physical priors that can only be learned from data.
This calls for latent 3D generative models, which have been shown to capture rich priors for 3D shapes~\cite{hunyuan3d2025hunyuan3d21,xiang2024structured,ke2024repurposing}, at least in the static case.
In this paper, we thus ask whether
\emph{such generative models can be extended to solve 4D reconstruction}, implicitly capturing cues such as symmetry and smoothness that can help recover the 3D mesh and its motion beyond what is directly visible in the video.
Recent contributions such as GVFD~\cite{zhang2025GVFD} have shown that this is a promising direction by introducing versions of 3D generators~\cite{xiang2024structured,hunyuan3d2025hunyuan3d21} that can handle dynamics.
However, they focus on generating good-looking images of the object
using a 3D Gaussian Splatting (3D-GS) representation~\cite{kerbl20233d},
and are less concerned with recovering accurate 3D shape and motion.

In this work, we introduce \methodname, a feed-forward model that learns to reconstruct \emph{accurate and complete dynamic 3D meshes} from a monocular RGB video.
We represent the object's shape as a 3D mesh extracted from the first frame of the input video, and then generate a \emph{deformation field} that displaces the surface of the mesh to represent the deformation of the object over time.
This representation is intuitive as it factorizes the 3D shape and motion.
By comparison, others~\cite{wang2025continuous,jiang2025geo4d,ShapeGen4D,ren2024l4gm} have approached 4D reconstruction by outputting \emph{independent} 3D reconstructions of each frame and thus do not explicitly model the object's motion.
Explicitly modeling motion is often crucial, for example, to texture the 3D object consistently over time.

We argue that, in order to accurately recover the object's dynamics,
they should be modeled holistically, from the beginning to the end of the video.
To make this feasible, we introduce a new Variational Auto-Encoder (VAE) to encode the mesh deformation in a compact latent space.
Prior work like Motion2VecSet~\cite{cao2024motion2vecsets} has proposed analogous encodings, but only for two frames at once.
In contrast, we show that a latent space representing the motion of the object throughout the whole sequence performs better.
To this end, we build a transformer~\cite{vaswani2017attention} encoder that uses spatial, temporal, and global attention modules~\cite{geyer2024tokenflow,wang2025vggt} in each block, capturing long-term correlations among points on the object.

We also argue for using motion-related information to guide VAE training,
which is a widely adopted strategy for static 3D latent spaces~\cite{li2025triposg,xiang2024structured,hunyuan3d22025tencent}.
In particular, we propose using the object's \emph{skeleton}, as this provides strong cues on the space of possible deformations of the object, explicitly coupling different physical points.
As shown in~\cref{fig:method_pipeline}, we leverage both skinning weights and bone information as additional inputs to the VAE transformer, which significantly improves the quality of the learned latent space.
Importantly, these are not needed at inference time.

Another challenge in monocular 4D mesh reconstruction is the lack of suitable 4D datasets for training and benchmarking.
Following recent works~\cite{ke2024repurposing,jiang2025geo4d,lu2025matrix3d,xu2025geometrycrafter}, we do not train our deformation model from scratch but start from pretrained \emph{static} 3D generators~\cite{hunyuan3d2025hunyuan3d21}.
These are trained on large datasets and help achieve robustness and generalization to a wide variety of object types.
However, we still need a 4D dataset with annotations for the motion of the 3D points (i.e., 3D tracks) for training and evaluation.
To this end, we build a synthetic dataset by filtering 3D animated assets from Objaverse~\cite{deitke2023objaverse}, which provides high-quality dynamic 3D meshes with skeletons.
We further extract dense 3D point correspondences across all frames.
For evaluation, we propose a new benchmark focusing on the quality of 3D shape and motion reconstruction rather than just visual quality.
To do so, we collect 50 animated 3D assets with significant object motion, high-quality textures, and no overlap with our training data.

To summarize, our contributions are as follows.
First, we propose a new framework, \methodname, for monocular 4D mesh reconstruction.
This model reconstructs the 3D shape and deformation of an object from a monocular video.
We build our model on top of a new VAE that encodes the object's deformation from the beginning to the end of the sequence in a compact latent space, suitable for latent diffusion.
Second, we propose a benchmark to assess monocular 4D mesh reconstruction models with a focus on 3D shape and motion reconstruction, which has often been neglected in favor of rendering quality.

\section{Related Work}%
\label{sec:related_work}

\subsection{Optimization-based 4D reconstruction}
Iterative or optimization-based methods reconstruct from monocular or multi-view videos by iteratively fitting a 4D representation to them~\cite{newcombe2015dynamicfusion,pumarola2021d,li2021neural,du2021neural,tretschk2021non,park2021nerfies,fridovich2023k,cao2023hexplane,li2023dynibar,Wu_2024_CVPR,yang2024deformable,yang2024real,som2024,wang2025freetimegs}.
With the advent of neural radiance fields (NeRFs)~\cite{mildenhall2020nerf} and 3D Gaussian Splatting (3D-GS)~\cite{kerbl20233d},
many time-dependent NeRFs~\cite{park2021nerfies,du2021neural,li2021neural,pumarola2021d,fridovich2023k,cao2023hexplane,li2023dynibar}
and dynamic 3D Gaussian Splatting methods~\cite{Wu_2024_CVPR,yang2024deformable,yang2024real,wang2025freetimegs,som2024} have been proposed for 4D reconstruction.
Although significant progress has been made, most approaches are evaluated on simple scenarios with quasi-static scenes~\cite{park2021nerfies,Wu_2024_CVPR} and relatively small datasets~\cite{yoon2020novel,gao2022monocular,li2022neural}.

To demonstrate effectiveness on diverse scenarios, recent works explore priors from large-scale pretrained models. One popular direction is to generate multi-view videos with pretrained video diffusion models~\cite{ho2022video,blattmann2023stable,blattmann2023align,bar2024lumiere} and then 
perform 4D reconstruction via per-scene optimization~\cite{xie2024sv4d,zeng2024stag4d,wu2025cat4d}.
Another line of work~\cite{jiang2024consistentd,zhang20244diffusion,dreamscene4d,ren2023dreamgaussian4d,li2024dreammesh4d} directly distills 4D knowledge from large-scale diffusion models using slow score-distillation sampling iterations~\cite{pooledreamfusion}.
However, these methods focus on “plausible'' novel view synthesis (NVS), rather than accurate geometry and tracking.
More recently, feed-forward 3D generators, such as Hunyuan3D~\cite{hunyuan3d22025tencent,yang2024hunyuan3d,hunyuan3d2025hunyuan3d21}, TripoSG~\cite{li2025triposg}, Trellis~\cite{xiang2024structured}, and Step1X-3D~\cite{li2025step1x}, have shown impressive results on diverse and complex scenarios, but they focus only on static 3D reconstruction.
Building upon these feed-forward 3D generators, V2M4~\cite{chen2025v2m4} reconstructs 4D meshes from monocular videos by generating per-frame 3D meshes independently, followed by an optimization step to ensure temporal coherence.
However, they still require per-scene optimization to achieve consistent 4D reconstruction, which is time-consuming and less flexible.
In contrast, our approach directly reconstructs temporally coherent 4D meshes along with tracking from monocular videos in a feed-forward manner.
It generalizes to diverse scenarios without per-scene optimization and is therefore much more efficient.

\subsection{Feed-forward 4D reconstruction}

Like our method, feed-forward 4D reconstruction directly infers a 4D representation from monocular videos in a single pass.
MonST3R~\cite{zhang24monst3r} effectively reconstructs dynamic scenes by adapting the \emph{static} 3D reconstruction produced by DUSt3R~\cite{wang2024dust3r} with dynamic scene supervision.
Numerous follow-ups take a similar path~\cite{st4rtrack2025,jin2024stereo4d,sucar2025dynamic}, adding functionalities like tracking.
However, they mainly work on pairwise frames or stereo videos.
To handle long monocular videos, Cut3R~\cite{wang2025continuous} introduces a memory bank to store spatial and temporal information, enabling continuous static and dynamic scene reconstruction from a stream of monocular images.
$\pi^3$~\cite{wang2025pi} further improves scalability by building a permutation-equivariant architecture on top of VGGT~\cite{wang2025vggt}. Beyond DUSt3R-based methods, Geo4D~\cite{jiang2025geo4d} leverages video generators~\cite{xing2024dynamicrafter} to infer 4D point maps from monocular videos.
4DGT~\cite{xu20254dgt} proposes a transformer model to predict dynamic 3D Gaussian representations~\cite{kerbl20233d} from monocular videos.
However, these approaches mainly focus on visible geometry reconstruction, whereas we address the more challenging task of complete 4D mesh reconstruction with tracking.
This is harder due to the need to infer the \emph{invisible} parts of surfaces and establish \emph{dense} correspondences over time.

Some recent works also explore complete 4D reconstruction from monocular videos in a feed-forward manner.
L4GM~\cite{ren2024l4gm} leverages a pretrained ImageDream~\cite{wang2023imagedream} to generate missing-view images, which are then used to train a feed-forward 4D Gaussian reconstructor.
GVFD~\cite{zhang2025GVFD} embeds the 4D mesh data into a Gaussian Variation Field using a compact latent space, and then trains a video-to-4D Gaussian model by inferring latent codes from monocular videos.
However, they still focus on producing plausible novel view synthesis, rather than aiming for accurate geometry.
More closely related to our task, concurrent work ShapeGen4D~\cite{ShapeGen4D} encodes a 4D mesh sequence into a sequence of latents,
and then learns a diffusion model to predict the latents directly, conditioned on monocular videos.
However, they align the latents only with a shared set of query points from the first frame and do not directly predict dense correspondences.
Instead, our \method is based on a novel spatio-temporal Transformer architecture operating directly on 4D meshes,
which can effectively capture temporal coherence and establish dense correspondences over time.
Moreover, compared to Motion2VecSets~\cite{cao2024motion2vecsets} which processes two meshes at a time, \method encodes long sequences of 4D meshes jointly, and directly predicts complete 4D meshes from monocular videos.

\section{Method}%
\label{sec:method}

\begin{figure*}[t]
    \centering
    \includegraphics[width=\textwidth]{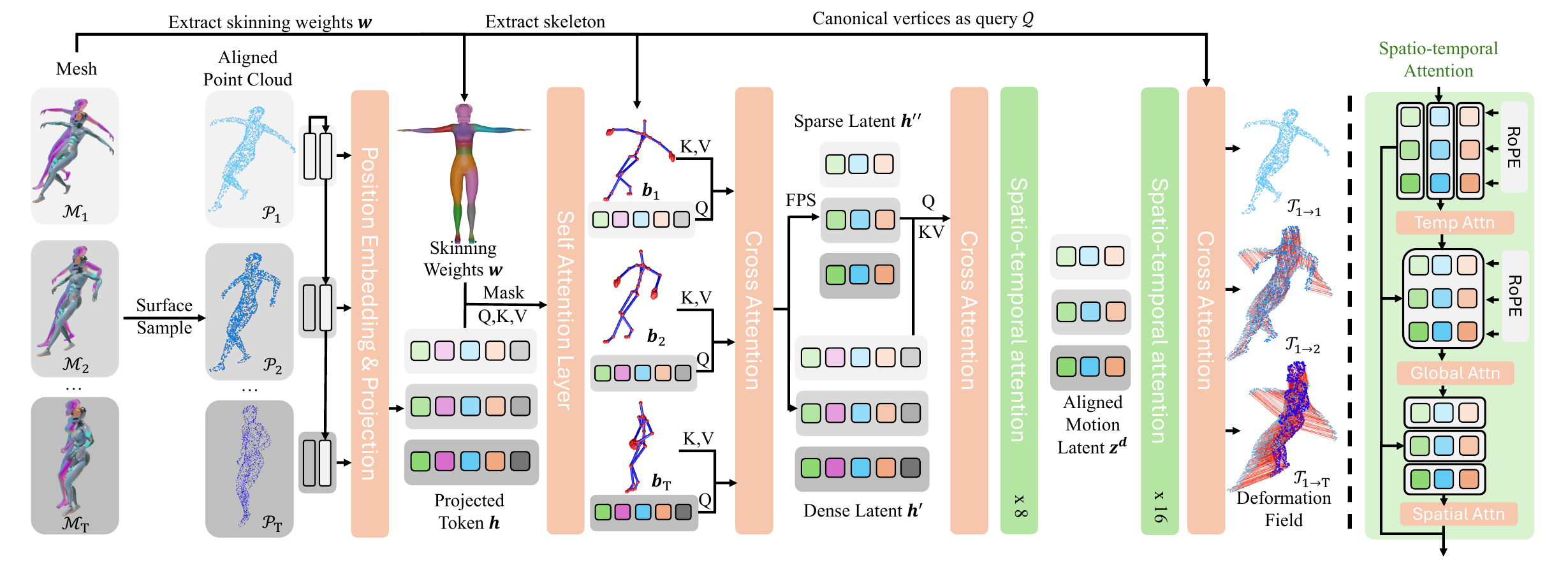}
     \vspace{-0.6cm}
    \caption{\textbf{Overall Deformation VAE pipeline.}
    (Left) Given a sequence of 3D meshes as input, we first uniformly sample a sequence of corresponding points.
    We inject the skeleton information by using masked self- and cross-attention. Then, a Farthest Point Sampling (FPS) at spatial dimension is performed to compress the latent, followed by 8 layers of spatio-temporal attention.
    The deformation field is decoded by layers of spatio-temporal attention, followed by a cross attention where canonical vertices serve as query points.
    (Right) Each of our spatio-temporal attention layers sequentially performs temporal attention, global attention, and spatial attention. For temporal and global attention, we additionally apply 1D RoPE~\cite{DBLP:journals/corr/abs-2104-09864} embedding on the temporal dimension.
    }
    \label{fig:method_pipeline}
\vspace{-0.2cm}
\end{figure*}

Given a monocular video of a moving object, our goal is to reconstruct its 3D shape and motion.
Formally, the video
$
\video = \{\image_t\}_{t=1}^{T}
$
is a sequence of $T$ RGB images
$
\image_t \in \real^{H \times W \times 3}.
$
The 3D shape of the object in the first frame of the video is captured by the \emph{3D mesh}
$
{\mesh}_1 = \langle \vertices_1, \faces_1 \rangle
$,
consisting of an array of 3D vertices 
$
\vertices_1 \in \real^{N_v \times 3}
$
and triangular faces
$
\faces_1 \in \mathbb{N}^{N_f \times 3}
$
indexing the vertices.
The motion of the object is captured by a dense \emph{deformation field}
$
{\deformation}_{1 \rightarrow t}
$
that specifies the displacement of all points on the mesh (including both vertices and faces) from time $1$ to time $t$.
In particular, we can use this deformation field to write the deformed mesh at time $t$ as
$
{\mesh}_t = 
\langle
     \vertices_1 + \deformation_{1 \rightarrow t}(\vertices_1), 
     \faces_1 
\rangle
$.

We cast our problem as learning a neural network $\Phi$ that, given the input video, outputs both the \emph{3D shape} and \emph{motion} of the object:
\begin{equation}
\label{eq:goal}
\Phi:
\video
\mapsto
\mesh_1, \{ {\deformation}_{1 \rightarrow t} \}_{t=1}^{T}.
\end{equation}
To ensure that the model generalizes well, especially given the limited availability of 4D training data, we build it on top of a pre-trained image-to-3D generator.
We opt for a 3D latent space model due to their robustness and ability to hallucinate complete objects even though only one aspect of them is visible in the image.

Our method has thus three main components.
First, it uses an off-the-shelf `backbone' network for image-to-3D reconstruction, which we introduce in \cref{sec:method-preliminaries}.
Its purpose is to reconstruct the mesh $\mesh_1$ from the first frame $\image_1$ of the video.
Second, it uses a new 4D Variational Auto-Encoder (VAE) to encode the deformation field $\deformation$ in a compact latent space (\cref{sec:method-vae}).
Third, it uses a generator that outputs the deformation latent code conditioned on the input video $\video$ and the mesh reference $\mesh_1$ (\cref{sec:method_diffuse}).

\subsection{Preliminaries: Latent 3D reconstruction model}%
\label{sec:method-preliminaries}

Following recent works on 3D diffusion~\cite{liu2023zero,ke2024repurposing,lu2025matrix3d,jiang2025geo4d}, we build \method on top of a pre-trained latent 3D reconstruction model for \emph{static} objects.
We use Hunyuan3D 2.1~\cite{hunyuan3d2025hunyuan3d21}, an excellent model learned from millions of 3D data samples, but any other similar model could be used instead.
\newcommand{\encoder}{\mathcal{E}}
\newcommand{\pointCloud}{\mathcal{P}}

The model uses a VAE to map the 3D mesh $\mesh$ to a compact latent code  $\latent^s$ based on the VecSet representation~\cite{zhang20233dshape2vecset}.
The \emph{encoder} $\latent^s=\encoder^s(\pointCloud;\normal)$ computes the latents by first uniformly sampling a point cloud $\pointCloud \in \real^{M \times 3}$ from the mesh $\mesh$, augmented with their normals $\normal \in \real^{M \times 3}$, and then applying a neural network to it.
Given the code $\latent^s$, a \emph{decoder} $\decoder^s(\queryPoints;\latent^s)$ recovers the 3D shape of the object by computing its signed distance function (SDF) in correspondence of a set of 3D grid query points $\queryPoints \in \real^{(H\times W\times D) \times 3}$.
These are finally converted to a triangle mesh via the marching cubes algorithm~\cite{lorensen1998marching}.

Given the VAE, 3D reconstruction amounts to learning a conditional denoising diffusion generator that can sample the latent code $\latent^s \sim p(\latent^s|\image)$ given the image $\image$.
This model is trained by minimizing the flow matching objective~\cite{lipman2023flow}: 
\begin{equation}
\label{eq:dm}
\min_{\parameters}
\mathbb{E}_{
    (\latent^s,\image),
    t,
    \noise^s \sim \mathcal{N}(\bm{0}, \bm{1})
}
\left \|
\velocity^s - \velocity^s_{\parameters} 
\left(\latent_t^s, t,\image\right)
\right \|_2^2,
\end{equation}
where
$
\latent_t^s = t \latent^s + (1 - t) \noise^s
$
is the noisy sample at timestep
$
t \sim \mathcal{U}(0, 1)
$,
and
$
\velocity^s = \latent^s - \noise
$
is the velocity field that moves the noisy sample $\latent_t^s$
towards the data $\latent^s$, and $\velocity^s_{\parameters}$ denotes the velocity prediction model where $\parameters$ are the model parameters.
Once trained, one randomly samples the noise from a Gaussian distribution and uses a first-order Euler ordinary differential equation to iteratively transfer the noise $\noise^s$ to the data $\hat{\latent}^s$.
Finally, the denoised shape latent $\hat{\latent}^s$ together with the 3D grid query points $\queryPoints$ are fed into the decoder to output the final mesh.

To obtain a textured mesh, a separate material generation model is used.
For this, the input image, rendered multi-view normal maps as well as canonical coordinate maps are taken as condition to generate a PBR texture.
Since texturing is orthogonal to our main contribution, we refer the reader to Hunyuan3D 2.1~\cite{hunyuan3d2025hunyuan3d21} for more details.
Note that this latent 3D reconstruction model is only used to reconstruct the \emph{static} mesh $\mesh_1$ from the first frame $\image_1$, which is used as the reference view for both geometry and texture.

\subsection{Deformation variational auto-encoder}%
\label{sec:method-vae}

In \cref{sec:method-preliminaries}, we have obtained the canonical mesh $\mesh_1=\langle \vertices_1, \faces_1 \rangle$ from the first frame $\image_1$ of the video using an \emph{off-the-shelf} model.
Our contribution is to reconstruct the deformation field $\deformation_{1 \rightarrow t}$ from $\mesh_1$ and the video $\video$ as a whole.

Just like the SDF defining the 3D shape of the object, the deformation field $\deformation_{1 \rightarrow t}$ is an infinite-dimensional object.
The key to our method is to learn a \emph{deformation VAE} that encodes the deformation field into a compact latent space.
The encoder
$
\latent^d \sim \encoder^d(\{\mesh_t\}_{t=1}^{T}; \normal, \skinning, \bone)
$
maps a sequence of meshes $\{\mesh_t\}_{t=1}^{T}$ to a latent code $\latent^d$, while the decoder takes this latent code and reconstructs the deformation field
$
\deformation_{1 \rightarrow t}(\vertices_1) = \decoder^d_t(\vertices_1;\latent^d)
$
in correspondence of the vertices $\vertices_1$.
The encoder also uses the normal vectors $\normal$, skinning weights $\skinning$, and bones $\bone$ to improve the encoding,
as we show in~\cref{fig:method_pipeline} and explain below.

\paragraph{Encoding points in time.}

Inspired by the design of the VAE in \cref{sec:method-preliminaries}, we build our deformation encoder by first extracting a 3D point cloud from the mesh.
However, in this case we are interested in encoding an \emph{animated mesh sequence} $\{\mesh_t\}_{t=1}^{T}$.
We assume that meshes in the sequence \emph{correspond}, meaning that the $i$-th vertex in each mesh corresponds to the $i$-th vertex in every other mesh.
We first extract a point cloud $\pointCloud_1$ by sampling mesh $\mesh_1$ as before.
We then express each 3D point in terms of $\vertices_1$ by finding its barycentric coordinates with respect to the face it belongs to, and determine its corresponding location at time $t$ by reconstructing it from vertices $\vertices_t$.
With this, we build a sequence of point clouds $\{\pointCloud_t\}_{t=1}^{T}$ that correspond to each other across time.

These points and corresponding normal vectors are combined and projected to a higher dimensionality:
\begin{equation}
\label{eq:input}
\features_{t} = f_l(
    \positionalEncoding(\pointCloud_1) \oplus \normal_1 \oplus \positionalEncoding(\pointCloud_t) \oplus \normal_t
),
\end{equation}
where $\positionalEncoding$ is positional embedding, $\oplus$ is the channel-wise concatenation operation (which makes sense because points correspond over time), and $f_l$ is a linear layer.
By pairing points at time 1 and $t$, we help the model figure out the motion of the object; by passing the normals, we further help it to determine the local motion.
Overall, we obtain a sequence of point features $\{\features_t\}_{t=1}^{T}$, where $\features_t \in \real^{M \times c}$ and $c$ is the feature dimension.

\paragraph{Injecting skeleton information.}

Just like with the normals $\normal$, we are free to use additional information to train a better deformation VAE\@, even if that information is not available at test time.
One of our contributions is to use skeleton information, captured by the skinning weights of the model, as privileged information at training time.
The \emph{skinning weights}
$
\skinning \in \real^{ M \times B_\text{max}}
$
represent the influence of each bone on each point, where $B_\text{max}$ is the maximum number of bones (we set $B_\text{max}=64$ and unused bones simply gets zero weights).
To encode the information in the skinning weights, we apply self-attention to the point features $\features_t$ with a bias that depends on skinning:
\begin{align}
\label{eq:weighted_attention}
\hat{\features}_t &= 
\operatorname{softmax} \left(
    \frac{\features_t {\features_t}^\top + M^s}{\sqrt{c}}
\right) \features_t 
+ \features_t \nonumber \\ 
& M^s  = \begin{cases}
 0      & \text{if}~ \skinning \skinning^\top >   \tau^s \\
 -\inf  & \text{if}~ \skinning \skinning^\top \le \tau^s
\end{cases},
\end{align}
where $c$ is the channel dimension and $M^s \in \real^{M \times M}$ is the attention masks which is calculated based on the similarity of the skinning weights and the threshold $\tau^s$.

In addition to biasing the self-attention layer, we incorporate the bone information via cross-attention.
Each bone is first represented by the position of its head and tail in each frame, captured by matrices $\bone_t^{h}, \bone_t^{t} \in \real^{B_\text{max} \times 3}$. Again, the unused bones are padded with zeros.
The bone parameters are then mapped to the same feature dimension as the point features $\features_t$ via a linear layer
$
\features^b_{t} 
= 
f^b_l(
    \positionalEncoding(\bone_1^{t}) \oplus
    \positionalEncoding(\bone_1^{h}) \oplus
    \positionalEncoding(\bone_t^{t}) \oplus
    \positionalEncoding(\bone_t^{h})
)$,
$\features^b_{t} \in \real^{B_\text{max} \times c}$.
Then, the point features attend these bone features via cross-attention:
\begin{align}
\features_t^\prime &= 
\operatorname{softmax} \left(
    \frac{\hat{\features}_t {\features_t^b}^\top + M^b}{\sqrt{c}}
\right) \features_t^b 
+ \hat{\features}_t \nonumber \\ 
& M^b = \begin{cases}
 0 & \text{if}~ \skinning > \tau^b \\
 -\inf  & \text{if}~ \skinning \le \tau^b
\end{cases},
\end{align}
where
$
M^b \in \real^{M \times B_\text{max}}
$
is the attention mask calculated based on the skinning weights.
In \cref{tab:ablation}, we show that incorporating the skeleton information in the encoder brings significant %
gain.
Note again that the skeleton information is only used in \emph{training} the deformation VAE to help learn a better motion prior, but is \emph{not} required during inference.

\paragraph{Spatio-temporal attention.}

Rather than modeling the motion of each frame independently~\cite{cao2024motion2vecsets}, we use a transformer~\cite{vaswani2017attention} with alternating attention layers~\cite{wang2025vggt}, including spatial attention, temporal attention, 
as well as global attention.
This allows the model to correlate the trajectories of different points on the mesh across the frames.

However, the computational cost of applying attention on all points across all frames is prohibitive.
Hence, we further downsample the points by applying Farthest Point Sampling (FPS), obtaining
$
\sparsePoints_t \in \real^{N \times c}
$
and performing cross-attention with 
$
\features^\prime_t \in \real^{M \times c}
$
to obtain a sparser set of latent vectors
$
\{\features^{0}_t\}_{t=1}^{T}
$,
$
\features^{0}_t \in \real^{N \times c}
$,
\begin{equation}
\label{eq:fps_attention}
\features^{0}_t =
\operatorname{softmax}\left(
    \frac{\sparsePoints_t {\features^\prime_t}^\top}{\sqrt{c}}
\right) \features^\prime_t,
\end{equation}
where $N \ll M$ and $c$ is the channel dimension.
Then, as shown in \cref{fig:method_pipeline} (right),
we apply spatial self-attention to the tokens $\features^{l}_t$ within each frame separately, temporal self-attention to the tokens across all frames but separate points, and global attention to all tokens $\cup_{t=1}^{T} \{\features^{l}_t\}$ across all frames jointly.
These attention layers are interleaved with MLP layers to further enhance the representation.

After $L=8$ layers of alternating attention,
an additional linear projection layer is applied to obtain the mean
$
\operatorname{E}(\features^{L}_t) \in \real^{N \times c_o}
$
and variance
$
\operatorname{Var}(\features^{L}_t) \in \real^{N \times c_o}
$
of the deformation latent distribution for each frame $t$, where $\features^{L}_t$ is the output of the last attention layer $L$ for frame $t$, and $c_o \ll c$.
Finally, the latent $\latent^d$ output by the encoder is sampled from
$
\mathcal{N}(
    \operatorname{E}(\features^{L}),
    \operatorname{Var}(\features^{L})
)
$.

\paragraph{Deformation field decoder.}

The decoder
$
\deformation_{1 \rightarrow t}(\vertices_1) = \decoder^d_t(\vertices_1;\latent^d)
$
recovers the deformation field in correspondence of the mesh vertices from the latent code $\latent^d$ output by the encoder.
First, we use a linear layer to project $\latent^d$ from dimension $c_o$ to a higher dimension $c$.
Then, we apply 16 layers of spatio-temporal attention blocks as described in~\cref{fig:method_pipeline} to further enhance the feature representation. Finally, the vertices $\vertices_1$ from the reference are used as queries to recover the deformation field
$
\deformation_{1 \rightarrow t}(\vertices_1)
$
via cross attention.

\paragraph{Training objective.}

The deformation VAE is trained by minimizing the following loss:
\begin{multline}
\label{eq:loss}
\mathcal{L}_\text{VAE} =
\sum_{t = 1}^{T}
\left \|
(\vertices_t - \vertices_1) - \decoder^d_t(\vertices_1; \latent^d)
\right \|_2^2 + \lambda L_\mathrm{KL},\\
\latent^d \sim 
    \encoder^d(\{\mesh_t\}_{t=1}^{T}; \normal, \skinning, \bone).
\end{multline}
$L_\mathrm{KL}$ is a KL divergence loss to regularize the latent space.
In practice, we evaluate this loss only for a random subset of vertices for efficiency.

\subsection{Deformation diffusion model}%
\label{sec:method_diffuse}

\begin{figure}[t]
\centering
\includegraphics[width=\linewidth]{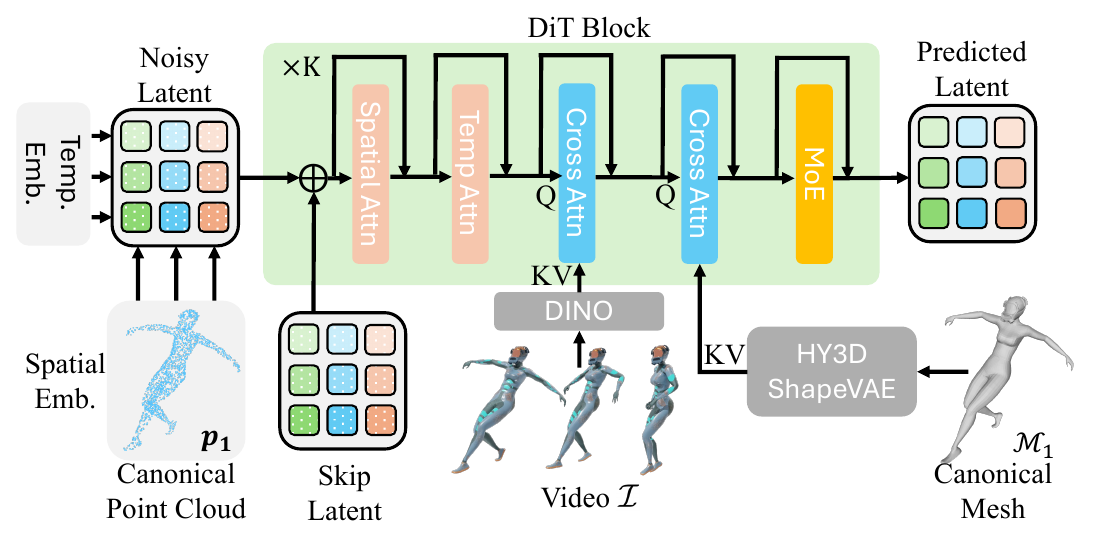}
\vspace{-0.3em}
\caption{\textbf{Overall deformation diffusion model pipeline.} We build it based on HY3D 2.1~\cite{hunyuan3d2025hunyuan3d21} shape diffusion model with additional spatial and temporal embedding as well as cross attention layer to condition the deformation field generation on the canonical mesh and input video.}%
\label{fig:method_diffusion}
\vspace{-0.3cm}
\end{figure}

As shown in \cref{fig:method_diffusion}, given the (reconstructed) canonical mesh $\mesh_1$ and the input video $\video$, we use a diffusion model to generate the deformation latent $\latent^d \sim p(\latent^d|\mesh_1, \video)$ conditioned on both mesh and video.
To build the deformation diffusion model, we extend the shape diffusion model in \cref{sec:method-preliminaries}:
it learns a velocity field $\velocity_{\parameters}^d$ with additional temporal embedding, and spatial embedding $\spatialemb_1 \in \real^{N \times 3}$ sampled from $\mesh_1$, as well as additional attention layers in each DiT block to incorporate the temporal information from the video and the shape information from the canonical mesh. The video feature is extracted from DINO-Giant~\cite{oquabdinov2} in a frame-wise manner and cross-attended by the corresponding latent.
Similar to GVFD~\cite{wu2024clusteringsdf}, the spatial embedding $\spatialemb_1$ gives a spatial awareness of the initial noise, which enhances the spatial consistency. During training, we use the positions of the FPS-sampled sparse features $\sparsePoints$ as a spatial embedding to formulate an aligned latent target and its embedding. During inference, we perform FPS sampling on the reconstructed canonical mesh to obtain a spatial embedding. Different from GVFD~\cite{wu2024clusteringsdf}, we %
also incorporate the temporal embedding for better temporal consistency and take advantage of the extracted high-dimensional shape feature $\boldsymbol{z}^s$ from the canonical mesh for better and detailed conditions. 
The training objective is similar to \cref{eq:dm}, where the velocity from noisy latent to data is taken as the target.

\section{Experiments}
\label{sec:experiments}

\begin{figure*}[t]
    \centering
    \includegraphics[width=\textwidth]{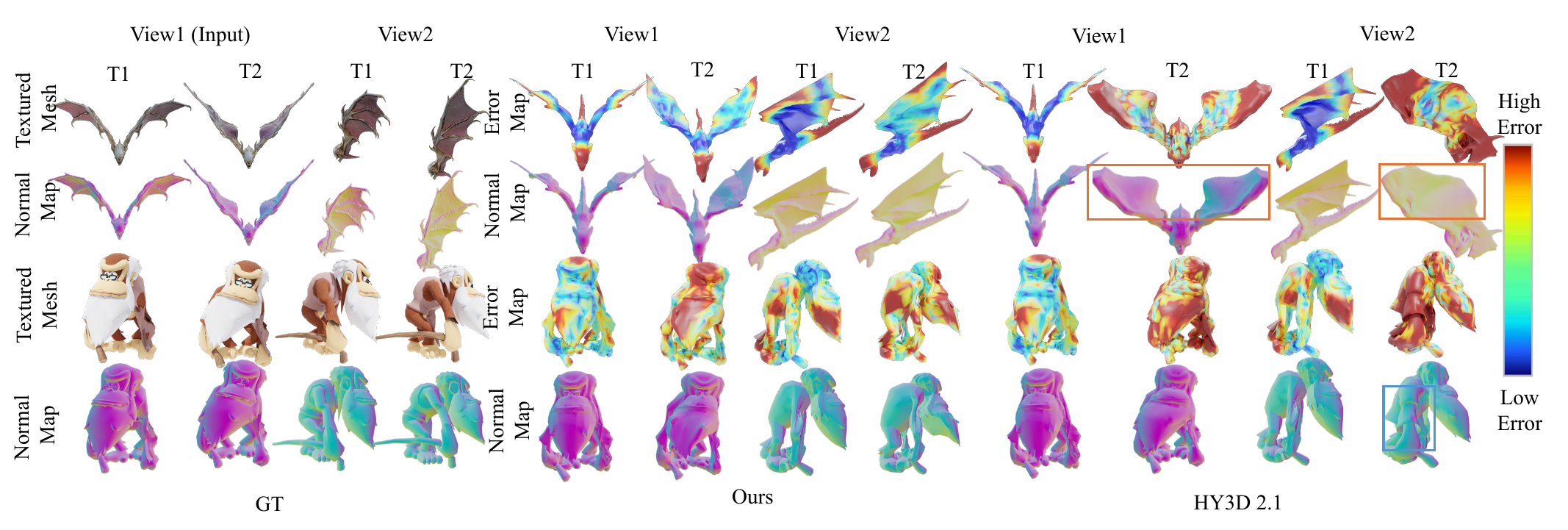}
     \vspace{-0.6cm}
    \caption{\textbf{Qualitative results on geometry reconstruction.} We show both the normal map and the error map (the bluer the better). HY3D 2.1~\cite{hunyuan3d2025hunyuan3d21} suffers from inaccurate \textcolor[RGB]{0,0,255}{\textbf{pose}} and \textcolor[RGB]{255,125,0}{\textbf{shape}} estimation due to the lack of temporal information. Thanks to the spatio-temporal attention, our method manages to reconstruct the mesh that follows the given input frames with accurate pose and similar shape.
    }
    \label{fig:rec_compare}
\end{figure*}

\tableRecon
\subsection{Experimental Settings}

\paragraph{Dataset.}

We start from the curated version of Objaverse-1.0~\cite{deitke2023objaverse} released by Diffusion4D~\cite{liang2024diffusion4d}, where objects animated with limited motion or large distortions are filtered out.
We extract the skeleton, skinning weights, and the sequence of meshes with corresponding vertices.
We then filter out objects with an excessively large number of vertices or bones,
resulting in approximately 9k instances.
Each instance is rendered as a frontal video with up to 100 frames.
For testing, we select a disjoint set of 50 mesh sequences with significant object motion and high-quality textures.
We render four fixed-view videos for each sequence at azimuth angles ${0^{\circ},90^{\circ},180^{\circ},270^{\circ}}$.
One view is used as the input; the remaining three are reserved for novel view synthesis (NVS) evaluation.

\paragraph{Baselines.}

We compare our method with three state-of-the-art latent 3D reconstruction models:
Hunyuan3D 2.1 (HY3D)~\cite{hunyuan3d2025hunyuan3d21},
L4GM~\cite{ren2024l4gm},
and GVFD~\cite{zhang2025GVFD}.
Since HY3D is an image-to-3D reconstruction method,
we run it frame by frame with shared sampled noise to improve temporal consistency.
For L4GM and GVFD, we take the centers of Gaussian primitives with opacity greater than 0.01 as a point cloud for geometry evaluation.
To align the reconstructed mesh with the ground-truth mesh,
we perform Coherent Point Drift (CPD)~\cite{DBLP:journals/corr/abs-0905-2635} to optimize scale and rigid transformation of the first-frame mesh. As shown in Tab.~\ref{tab:recon} and Tab.~\ref{tab:nvs}, entries with the \textit{Aligned} annotation indicate that the aligned canonical shape is used as input to our deformation diffusion model, whereas entries without this annotation indicate that alignment is performed only immediately before evaluation.

\paragraph{Metrics.}

For frame-wise \emph{geometry} evaluation we report volumetric IoU, point-to-surface distance (P2S), and Chamfer distance.
For \emph{tracking}, we measure the Euclidean distance between corresponding points (nearest neighbors on the first frame) on the predicted and ground-truth meshes ($\ell_2$-Corr).
For \emph{novel view synthesis (NVS)}~\cite{zheng2024free3d,chen2024mvsplat},
we report Peak Signal-to-Noise Ratio (PSNR), Structural Similarity Index Measure (SSIM), Learned Perceptual Image Patch Similarity (LPIPS), and CLIP similarity.
We also report FVD~\cite{unterthiner2019fvd} to assess the temporal consistency of generated novel-view videos.

\subsection{Geometry and tracking evaluation}

\begin{figure*}[t]
    \centering
     \vspace{-0.5cm}
    \includegraphics[width=\textwidth]{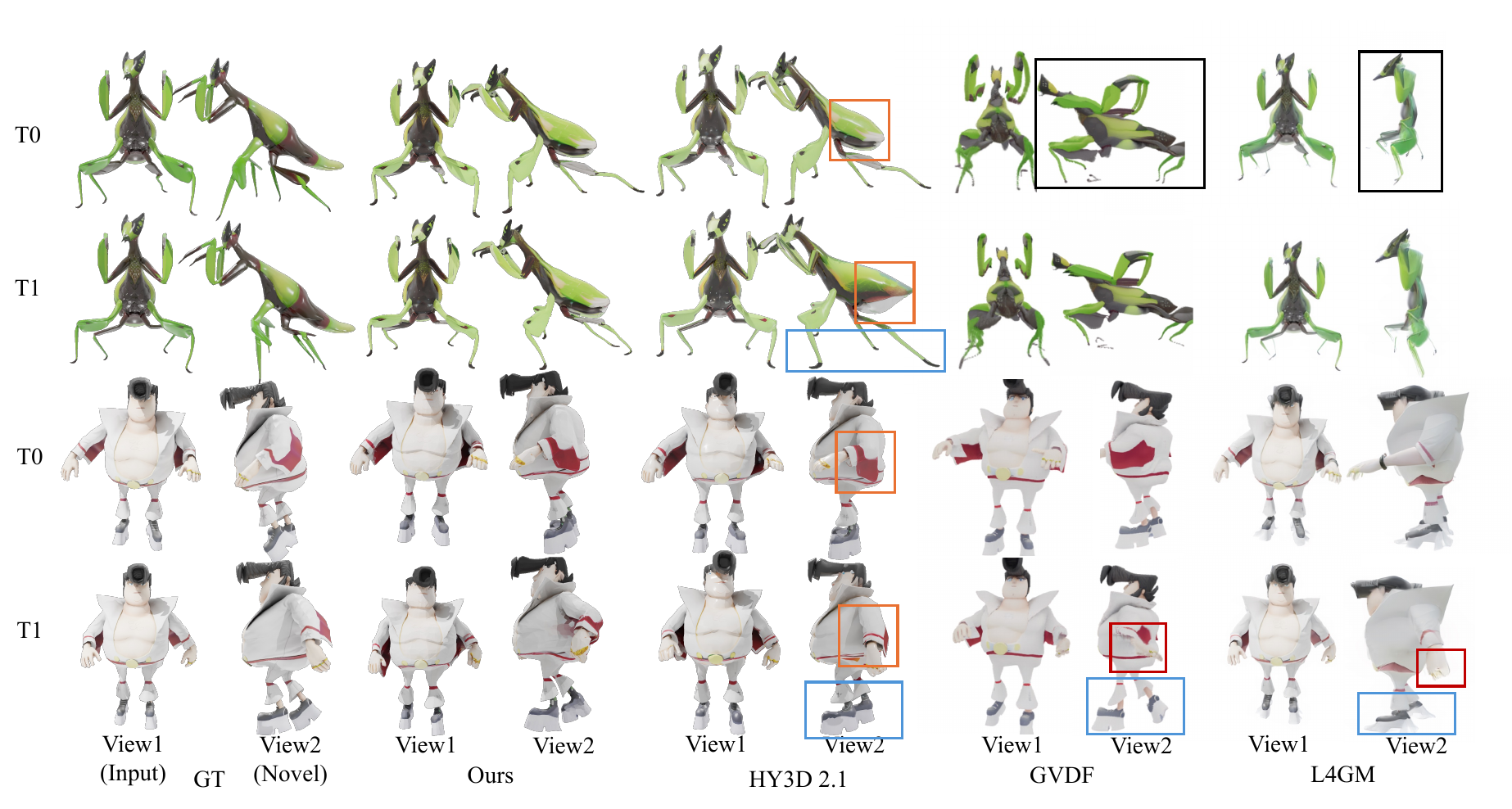}
     \vspace{-0.5cm}
    \caption{\textbf{Qualitative results on novel view synthesis.} 
    All the state-of-the-art methods suffer from  \textcolor[RGB]{0,0,255}{\textbf{inaccurate pose estimation}},
    either due to lack of temporal attention (HY3D~\cite{hunyuan3d2025hunyuan3d21}) or neglect the importance of geometric supervision (GVDF~\cite{zhang2025GVFD},  L4GM~\cite{ren2024l4gm}).
    3D-GS based methods occasionally exhibit \textcolor[RGB]{255,0,0}{\textbf{ghost artifacts}} because they lack topology constraints during deformation,
    while the frame-wise reconstruction method produce \textcolor[RGB]{255,125,0}{\textbf{inconsistent shape and texture}}.
    Moreover, by leveraging a large reconstruction method, we avoid predicting extremely \textbf{incorrect canonical mesh}.
    Thanks to the skeleton information and spatio-temporal attention, \method is able to reconstruct accurate pose and geometry, and produces temporally consistent novel view video.
    }
    \label{fig:nvs_compare}
\vspace{-0.5cm}
\end{figure*}

As shown in \cref{fig:rec_compare}, we qualitatively compare our method with frame-wise HY3D inference~\cite{hunyuan3d2025hunyuan3d21}.
Due to the lack of temporal attention across frames,
HY3D suffers from inaccurate pose (incorrect pose of the chimpanzee's arm in the \textcolor[RGB]{0,0,255}{\textbf{blue}} box) and shape (incorrect bat wing shape in the \textcolor[RGB]{255,125,0}{\textbf{orange}} box).
This is confirmed by the quantitative results in \cref{tab:recon},
where \method achieves state-of-the-art reconstruction and tracking performance by leveraging information across the whole sequence.
3D-GS-based methods do not inherently focus on geometry or dense tracking,
which is also reflected in the ghost artifacts (see \cref{fig:nvs_compare}) discussed next.

\subsection{Novel view synthesis evaluation}

In \cref{fig:nvs_compare} we qualitatively assess NVS results on both the input view and a novel view at different time steps.
Typical errors fall into four categories:

\paragraph{Inaccurate pose estimation.}

As shown in the \textcolor[RGB]{0,0,255}{\textbf{blue}} boxes of \cref{fig:nvs_compare},
HY3D predicts inaccurate poses (e.g., human legs, mantis limb).
In contrast, our method predicts pixel-aligned pose estimates via temporal and global attention,
yielding better PSNR, SSIM, and LPIPS (see \cref{tab:nvs}).

\paragraph{Inconsistent texture.}

Although using shared noise improves HY3D frame-to-frame consistency, it still exhibits jittery geometry and texture flicker (see the \textcolor[RGB]{255,125,0}{\textbf{orange}} boxes).
Our model produces consistent texture and geometry thanks to dense correspondences modeled by the spatio-temporal attention, leading to a lower FVD (\cref{tab:nvs}).

\paragraph{Incorrect canonical mesh.}

Our %
\dlt{method} builds upon a large-scale 3D reconstruction model, which is able to produce roughly good canonical shape for various objects. On the contrary, GVFD and L4GM often fail to recover an accurate canonical shape,
a prerequisite for high-quality 4D reconstruction (\textbf{black} boxes in \cref{fig:nvs_compare}).

\paragraph{Artifacts of deformed 3D-GS.}
Prior methods that generate 3D-GS representations purely rely on photometric losses without explicit geometry and topology constraints. They suffer from ghost artifacts under large motion
(\textcolor[RGB]{255,0,0}{\textbf{red}} box in \cref{fig:nvs_compare}).

Our texture map is generated from only the first input frame,
so our method yields a lower CLIP score compared to HY3D which generates a new texture for each frame.
However, thanks to the reconstructed deformation field, we estimate more accurate motion and alignment,
achieving state-of-the-art results on all other metrics (\cref{tab:nvs}).
\tableNVS

\begin{figure}[t]
\centering
\includegraphics[width=\linewidth]{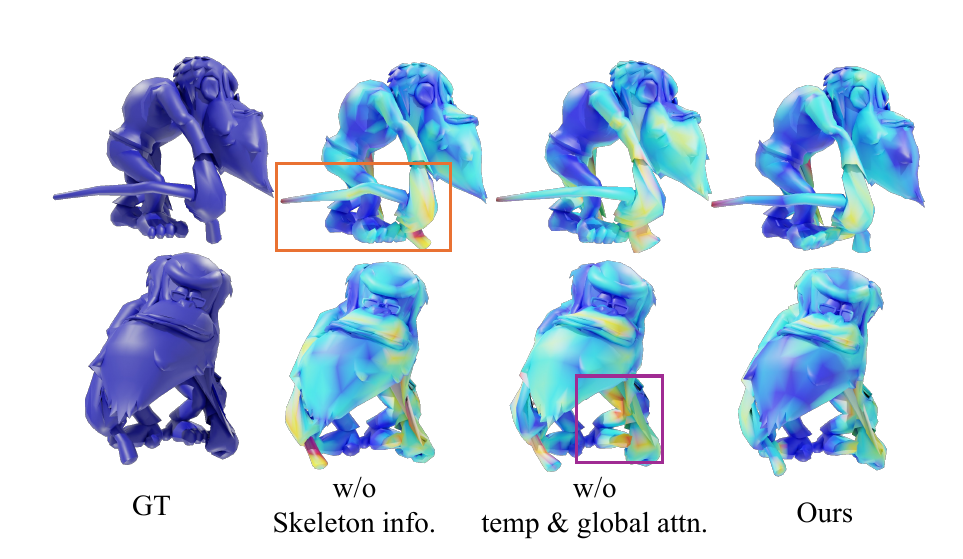}
\vspace{-0.6cm}
\caption{\textbf{Ablating key components of the deformation VAE.}
We visualize the error map of the chamfer distance,
where blue indicates lower error (better reconstruction quality).
Injecting skeleton information helps the model better capture rigid deformation,
while spatio-temporal fusion effectively reduces jittering effects.
}%
\label{fig:ablation}
\vspace{-0.4cm}
\end{figure}

\subsection{Ablation study}

\tableAblation

We ablate key design choices of the deformation VAE\@.
We train two variants: one without skeleton information and one without temporal and global attention.
At test time we use the ground-truth first-frame mesh as the canonical mesh and evaluate the reconstructed deformation field.
As shown in the \textcolor[RGB]{255,125,0}{\textbf{orange}} box of \cref{fig:ablation}, removing skeleton information impairs rigid transformations, resulting in a twisted stick.
As shown in the \textcolor[RGB]{157,44,146}{\textbf{purple}} box, removing temporal and global attention causes jittery motion and larger errors near the feet.
These observations align with the quantitative results in \cref{tab:ablation}, confirming the effectiveness of our VAE design.

\section{Conclusion}%
\label{sec:conclusion}

We introduced \method, a feed-forward approach for 4D mesh reconstruction from monocular videos.
Starting from a latent 3D reconstruction model pre-trained on a large collection of static objects, we add a new VAE that encodes the object deformation in a compact latent space, a method to use skeleton information to supervise this VAE, and a new diffusion model built on these components.
With these, \method is able to predict the full 3D shape of the object as well as its deformation, tracking vertices throughout the entire video sequence. 

On the Objaverse benchmark, \method achieves state-of-the-art reconstruction quality for geometry, correspondence, and novel-view synthesis, while reducing temporal artifacts. 
Our ablations highlight the value of our contributions, %
including using skeletal cues and spatio-temporal attention in the VAE architecture.
Limitations include reliance on a high-quality canonical mesh and skeletons for training, the inability to represent topological changes in the mesh, and difficulty in reconstructing extremely non-rigid objects.
We will release code, models, benchmark splits as well as the whole evaluation framework.

\paragraph*{Acknowledgments.}

The authors of this work were supported by Clarendon Scholarship, NTU SUG-NAP, NRF-NRFF17-2025-0009, ERC 101001212-UNION, and EPSRC EP/Z001811/1 SYN3D.

{
    \small
    \bibliographystyle{ieeenat_fullname}
    \bibliography{chuanxia_general,chuanxia_specific,main}
}

\clearpage
\maketitlesupplementary

In this \textbf{supplementary document}, we provide additional materials to supplement our main submission. In the \textbf{supplementary video}, we show more visual results using our method. 
The \textbf{code, models, benchmark splits, and evaluation framework} will be made publicly available for research purposes.

\section{Implementation Details}
\label{sec:rationale}

\subsection{Training details}

\noindent \textbf{Deformation VAE.}
The initial weights of our deformation VAE is loaded from HunYuan3D 2.1 shape VAE.
All the last projection layers of the additional introduced modules are zero initialized, \ie the skeleton injection layer and the spatio-temporal attention layer.
The Deformation VAE is trained using AdamW with a learning rate of $1 \times 10^{-5}$ and a batch size of 80.
We set $M=2048$ for the initial sampled aligned point cloud, and $N=256$ for the number of point cloud after Farthest Point Sampling. The hidden dimension $c=1024$ is set for the attention operation, and $c_0 = 64$ is set for the latent space. 
The weight of the KL divergence loss is set to $\lambda=5 \times 10^{-5}$.
Due to the limited computational resources, we only train our model with the frame number $T=6$.
However, thanks to our mesh representation, during animation, it is commonly to model only the key frames and do the shape interpolation between them, instead of training an interpolation model in L4GM~\cite{ren2024l4gm} specialized for 3D-GS.
In our supplementary video, we do the one frame shape interpolation between two key frames, resulting in a total 11 frames per sequence. During training, for each sample, we select 6 frames from the sequence, with the sampling stride randomly chosen from $[1, 2, 3, 4]$ to allow our model to adapt to input videos with various frame rates.
Training is conducted on 4 NVIDIA H100 GPUs with a total training time of approximately 5 days. 

\noindent \textbf{Deformation diffusion.}
Our deformation diffusion model is initialized with the weights of HunYuan3D 2.1~\cite{hunyuan3d2025hunyuan3d21} diffusion model and trained using AdamW with a learning rate of $1 \times 10^{-5}$ and a batch size of 80.
Similarly, we perform zero initialization to newly introduced modules, including canonical shape condition layer and temporal attention layer.
The dimension of the latent feature from the HY3D ShapeVAE is $256 \times 64$.
Training is conducted on 4 NVIDIA H100 GPUs with a total training time of approximately one week. 

\subsection{Inference details}
During inference,
given a monocular video of a moving object,
we first segment the foreground moving object using a pre-trained model,
and then resize to the same ratio as the training (90\%) for condition.
For the canonical shape reconstruction, we follow the same inference setting as HunYuan3D 2.1~\cite{hunyuan3d2025hunyuan3d21},
but using only 1 view (the first frame) as input.
For the deformation diffusion model,
we perform 50 steps of first-order Euler ordinary differential equation (ODE) to transform the sampled noise to the desire deformation latent,
conditioned on the canonical shape latent and the image features from all input frames,
Once we obtain the deformation latent, we reconstruct the per-vertex deformation field using the deformation decoder.

\section{Additional Analysis}
\subsection{Ablation study for CFG}
\tableCFG
We perform classifier-free guidance (CFG) with the guidance weights of $5$. As shown in Tab.~\ref{tab:cfg}, CFG does not help reconstruction quality.
This observation aligns with other video-based diffusion reconstruction methods~\cite{jiang2025geo4d}.

\subsection{Ablation study for pretrained diffusion model}
\tablePretrain
As claimed in~\cref{sec:intro},
due to the limited size of existing 4D reconstruction datasets,
we build our deformation diffusion model upon a pre-trained large-scale 3D generator, \ie HunYuan3D 2.1~\cite{hunyuan3d2025hunyuan3d21}.
To validate this design choice, we conduct an ablation study.
Specifically, we train two versions of our deformation diffusion model for the same number of iterations (3 days), one with HY3D pre-training weights and one without.
The results are reported in Tab.~\ref{tab:pretrained}.
As can be seen, the model with HY3D pre-training weights outperforms the one without by a large margin in terms of all metrics, demonstrating that the pre-training weights obtained from large-scale 3D reconstruction dataset indeed benefit our deformation diffusion model.

\begin{figure*}[t]
    \centering
    \includegraphics[width=\textwidth]{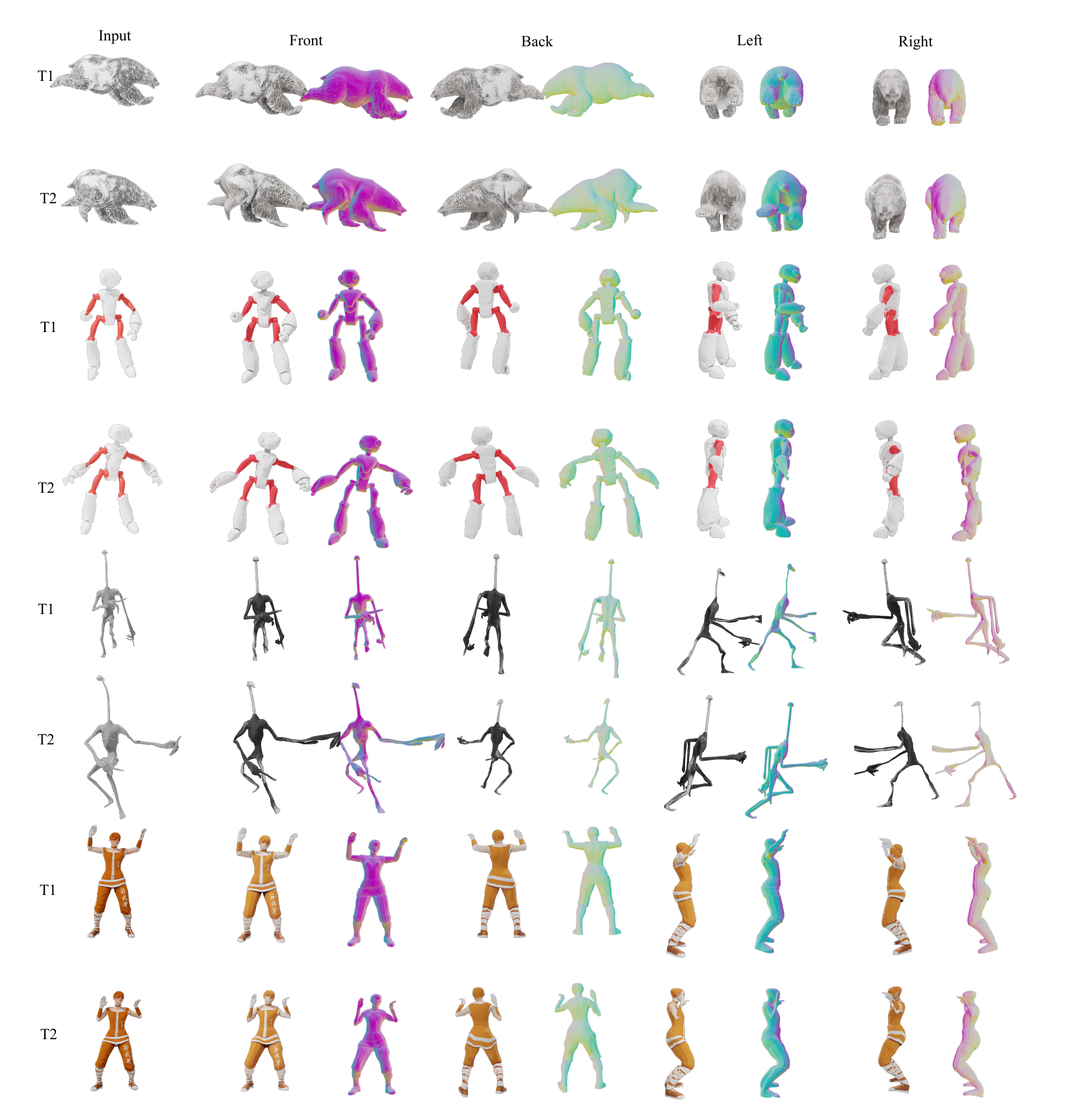}
     \vspace{-0.6cm}
    \caption{\textbf{More visualization results.} The left column is two frames sampled from the input video, the others are the corresponding reconstruction results from 4 different views.
    }
    \label{fig:vis}
\vspace{-0.3cm}
\end{figure*}

\section{Visualization}
As shown in Fig.~\ref{fig:vis}, our method can generalize well on various objects and motions.

\begin{figure}[t]
\centering
\includegraphics[width=\linewidth]{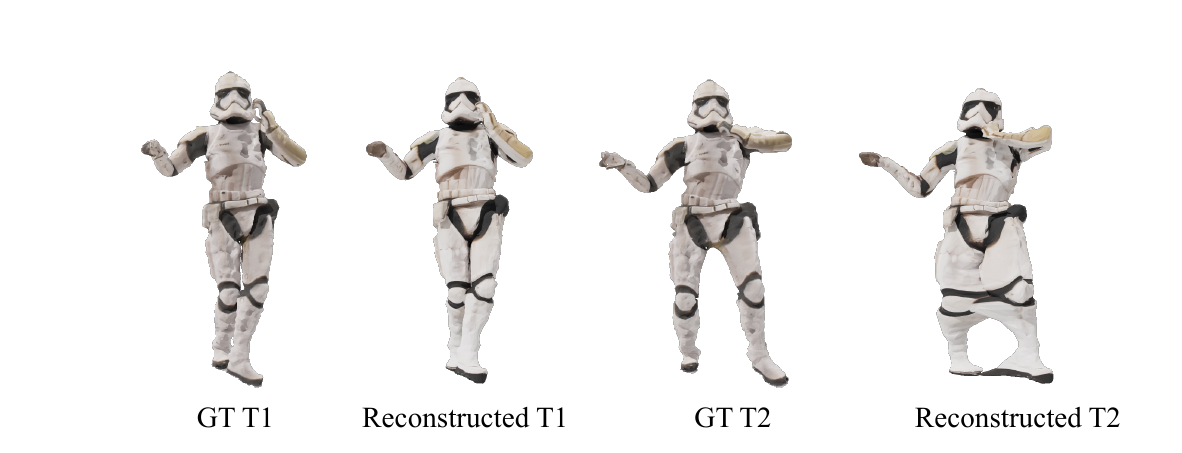}
\vspace{-0.0cm}
\caption{\textbf{Failure case of Mesh4D.}
}%
\label{fig:limitation}
\vspace{-0.4cm}
\end{figure}

\section{Limitations}
\label{sec:limitation}
Although our method performs well and generalizes to a wide range of objects and animations, it can fail when the topology changes a lot during animation, or the method fails to reconstruct correct topology or shape for canonical mesh.
As shown in~\cref{fig:limitation}, when the 3D reconstruction model fails to predict separate legs in the first frame, even if our model predicts the deformation field for the following frames, the topology of the subsequent mesh remains unchanged, leading to the incorrect 4D reconstruction.
However, this problem can be easily solved by choosing a different frame to reconstruct the canonical shape and perform a backward and forward deformation field reconstruction.
However, how to choose a good reference frame for the canonical mesh is orthogonal to our contribution and beyond the scoop of this paper.

\end{document}